# Hybrid Life: Integrating Biological, Artificial, and Cognitive Systems

Manuel Baltieri[1,2 †], Hiroyuki Iizuka[3,4], Olaf Witkowski[5], Lana Sinapayen[6], Keisuke Suzuki[4]

[1] Araya, Tokyo, Japan
[2] Department of Informatics, University of Sussex, Brighton, UK
[3] Faculty of Information Science and Technology, Hokkaido University, Sapporo, Japan
[4] Center for Human Nature, Artificial Intelligence and Neuroscience (CHAIN), Hokkaido University, Sapporo, Japan
[5] Cross Labs, Cross Compass, Kyoto, Japan
[6] Sony Computer Science Laboratories, Kyoto, Japan

[†] Corresponding author - manuel_baltieri@araya.org

## Abstract

Artificial life is a research field studying what processes and properties define life, based on a multidisciplinary approach spanning the physical, natural and computational sciences. Artificial life aims to foster a comprehensive study of life beyond "life as we know it" and towards "life as it could be", with theoretical, synthetic and empirical models of the fundamental properties of living systems. While still a relatively young field, artificial life has flourished as an environment for researchers with different backgrounds, welcoming ideas and contributions from a wide range of subjects. *Hybrid Life* is an attempt to bring attention to some of the most recent developments within the artificial life community, rooted in more traditional artificial life studies but looking at new challenges emerging from interactions with other fields. In particular, Hybrid Life focuses on three complementary themes: 1) theories of systems and agents, 2) hybrid augmentation, with augmented architectures combining living and artificial systems, and 3) hybrid interactions among artificial and biological systems. After discussing some of the major sources of inspiration for these themes, we will focus on an overview of the works that appeared in Hybrid Life special sessions, hosted by the annual Artificial Life Conference between 2018 and 2022.

## 1. INTRODUCTION

Artificial life (ALife) has emerged in the last few decades as a field studying "life as it could be" (Langton, 1989). Starting from the origins of living systems known on Earth, ALife's main goal is a general definition of life, its properties and functions, that captures a series of fundamental principles beyond "life as we know it". ALife research is often classified based on the type of substrates used in different experiments. These substrates include: simulation-based work, hardware implementations, and chemical syntheses of life-like systems, going by the name of soft, hard and wet artificial life respectively (Bedau, 2003). Prototypical examples of ALife research in these areas include simulations for the emergence of (digital) complexity and abiogenesis (Ofria & Wilke, 2004; Beer, 2014; Agmon et al., 2016; Chan, 2020), evolutionary robotics (Harvey et al., 2005; Kriegman et al., 2020), and chemical compounds displaying forms of homeostasis, self-replication and evolution akin to living systems (Hanczyc & Ikegami, 2010; Čejková et al., 2017).

ALife's roots also run deep into cybernetics (Wiener, 1948; Ashby, 1957), connecting this field to formal studies of intelligence and cognition. Owing to this, modern ALife research include for instance work characterising agency (Barandiaran et al., 2009; Beer, 2014; Biehl, 2017), autonomy (Di Paolo, 2005; Di Paolo & Iizuka, 2008), autopoiesis (Beer, 2004; 2015; 2020; Suzuki & Ikegami, 2009) and the (enactive) mind (Froese & Ziemke, 2009; as inspired by Varela et al., 1991), as well as minimal cognition (Beer, 1995; Harvey et al., 2005; McGregor et al., 2015), and learning and adaptation in artificial and biological systems (Sinapayen et al., 2017; Baltieri & Buckley, 2017).

In recent years, the boundaries of ALife work have broadened to include topics traditionally found in other fields, giving rise to a process of cross-pollination of ideas across disciplines often based on the "*something* as it could be" mantra, including cognition (McGregor et al., 2007), machines (Bongard & Levin, 2021), chemistry (Banzhaf & Yamamoto, 2015) and the mind (Froese et al., 2012) as they could be. More concretely, ALife has come to embrace and interact with fields such as policy making (Penn, 2016), psychology (Sinapayen et al., 2021) and engineering (Suzuki et al., 2017; Suzuki et al., 2019a) among others. This overview brings forward a number of directions that ALife researchers, including the authors, have explored, and that are increasingly indistinguishable from other fields, including psychology, human- and brain-machine interaction, information and control theory, virtual reality technologies, and others. We dub this area "Hybrid Life".

In Section 2 we will focus on a systematic characterisation of Hybrid Life and its three main themes: theories of systems, hybrid augmentation and hybrid interaction. This will provide the basis for our research agenda in this area, providing an incomplete list of research topics and results that contributed to the definition of Hybrid Life. Section 3 will then provide a report of the contents discussed and the proceedings published as part of the annual International Conference on Artificial Life in the period 2018-2022, where Hybrid Life appeared as one of the special sessions connecting ALife to other fields. In our concluding remarks (Section 4), we will look at Hybrid Life's future developments.

## 2. HYBRID LIFE

Hybrid Life focuses on three areas of research that have recently emerged within ALife and that are at least partly inspired by *systems theory*. Systems theory is an interdisciplinary research field at the intersection of control theory, biology, complexity science and psychology, among others, concerned with the characterisation and understanding of systems, their behaviours and interactions (Von Bertalanffy, 1972; Skyttner, 2005). In Hybrid Life, our focus is particularly on:

- **Theories of systems and agents**, theoretical frameworks focusing on formal definitions of systems and agents that can take into account different features of biological, cognitive and artificial systems,
- **Hybrid augmentation**, looking at how autonomous systems' capabilities can be extended via the integration with non-autonomous parts,
- **Hybrid interactions**, describing how autonomous systems of different kinds can interact to generate complex, or perhaps "emergent" behavioural patterns, and how these patterns of interaction affect the independent autonomous systems.

### 2.1 Theories of systems and agents

Studies of systems have often been tackled independently in different research areas: engineering, computer science and physics for the artificial, biology for the living and neuroscience/psychology for the cognitive realms. Agents, informally defined as *autonomous* systems *acting* on their environment to fulfil their *goals*, have been either treated as "mere" systems, discounting the role of autonomy, actions and goals, or taken for granted based on folk-psychological understandings of these concepts (for some exceptions to both cases, see for instance Beer, 1995; Barandiaran et al., 2009; Biehl, 2017).

For example, the emergence of synthetic agents in ALife is often abstracted away from biological and physical substrates to focus on their individuality, adaptivity and complex behaviours (Suzuki & Ikegami, 2009; Beer, 2014, 2020; Agmon et al. 2016; Chan, 2020). While crucial for exploring agency, life, cognition, etc. *as they could be*, this may discount substrate-specific properties that might be necessary for the definition of general systems. On the other hand, in standard approaches to life sciences, living systems are usually studied by focusing on properties that particular chemical substrates appear to induce, such as self-organisation, homeostasis, metabolism and self-replication (see for example Lanier & Williams, 2017; Preiner et al., 2020). This approach may shed light on the importance of biological and biochemical properties of life as we know it, but it could also just lead us to misunderstandings about the fundamental building blocks of life, missing the forest for the trees. Cognition is on the other hand often seen as orthogonal to a theory of systems, and too complex to relate to agency. Ideas such as embodiment and situatedness (Brooks, 1991; Newen et al., 2018), the mind-life continuity thesis (Maturana & Varela, 1980; Varela et al., 1991) and minimal cognition (Beer, 1995; Harvey et al., 2005) however challenge this view by noticing that studies of agents, biological and artificial ones might be intrinsically tied to cognition, albeit often in its minimal forms.

To bring back the focus to approaches that encompass different classes of systems, and to consider the role of actions and goals for agents, we introduced "theories of systems" as a theme in Hybrid Life, looking at it as a *trans*-disciplinary, independent research field about studies of systems and agents. A paradigmatic historical example of this trans-disciplinary approach is cybernetics (Wiener, 1948; Ashby, 1957), a field that over decades of development and various reboots, brought together ideas from control theory and engineering, dynamical systems, cell and neurobiology to study complex and cognitive behaviour of both biological (Maturana & Varela, 1980; Varela et al., 1991) and artificial (Walter, 1950; Beer, 1984) systems. Based on cybernetic arguments on the importance of interactions between agents and their environments, Beer (1995) later built a mathematical treatment of some general aspects of agency based on interacting dynamical systems representing brain, body and environment under rather general assumptions. Beer (2004) also introduced further formalisations of a core concept developed under the influence of "second-order" cybernetics, autopoiesis. In particular, Beer (2004) defined a precise mathematical account of some features of autopoiesis for general systems, starting with "gliders", spatio-temporal moving patterns within the Game of Life cellular automaton.

Expanding on Beer (2004)'s work, Biehl et al. (2016) built a mathematical theory of agency on notions of spatio-temporal patterns using probabilistic modelling (Bayesian networks) and a measure of individuality based on integrated information theory (IIT) (Oizumi et al., 2014). IIT was originally proposed as a way to mathematically formalise consciousness following a definition of "integrated information". Integrated information quantifies the so-called "intrinsic irreducibility" of a system, i.e., how much cause-effect power can be found at a specific scale of a system (the whole) that cannot be reduced to smaller partitions (the parts). More recently however, IIT has also been used in other contexts: for instance, Albantakis & Tononi (2015) adopted IIT to define aspects of agency in the spatial domain, while Albantakis et al. (2021) further dove into the temporal continuity of agents. Probabilistic modelling in the form of Bayesian networks and related approaches is also adopted by proponents of the free energy principle (FEP), initially described as a unifying principle of brain functioning (Friston, 2010), and more recently providing a perspective on systems, or *things* (Friston, 2019; Friston et al., 2022). Under the FEP, a system (or a thing) is something that appears to minimise variational free energy, i.e., to perform (approximate) Bayesian inference on its environment. Interestingly, this can happen for different partitions of a system and at different scales at the same time. While this can lead to a theory where everything might be a *system* without any further constraints (Bruineberg et al., 2021), unlike other approaches, such as standard IIT implementations

(Oizumi et al., 2014), the FEP embodies crucial aspects of "being a system" (or agent) at multiple scales: the possibility that societies, oneself and the cells within one's body can all be seen as agents at the same time.

The probabilistic approach has also led to proposals such as Krakauer et al. (2020) where we find definitions of system-environment partitions relying on ideas of autonomy, using either predictive information[1] to measure the self-predictability of the state of a candidate system (the more self-predictability, the more agency), or predictive information conditioned on the states of the environment, i.e., how much more the past of a system can tell us about its future beyond what's contained in observations about the environment. Kolchinsky & Wolpert (2018) expand on some of these ideas to further add a constraint of causal necessity by "scrambling" information between system and environment to define syntactic and semantic information, proxies for correlational and causal information respectively. Following this approach, Kolchinsky & Wolpert (2018) provide measures such as "stored semantic information", the mutual information between a system and its environment with scrambled probabilities at their initial condition, and "observed semantic information" as a scrambled transfer entropy to measure how observations of an environment help predict the future of a system. Kolchinsky & Wolpert (2018) then conjectures that autonomous agents will generally display high levels of both kinds of semantic information, with different attributes derived from a focus on stored or observed semantic information.

More recently, further mathematical explorations of the idea of systems have been proposed in "categorical cybernetics" (Capucci et al., 2021), using the language of category theory (see also Rosen, 1991). For instance, Fong et al. (2021) present a treatment describing bidirectional behavioural constraints between the "whole" and the "parts" of a system, generalising some aspects of proposals such as IIT and FEP. Myers (2021) introduces a similar treatment but in a different categorical generalisation, i.e., using a definition of a double category of open dynamical systems, where behaviours of the parts and their functional composition into behaviours of the whole is a complementary aspect of the structural composition of parts into wholes. Virgo et al. (2021) and Biehl & Virgo (2022) instead focus on a general notion of agents in terms of systems whose state evolution can be interpreted as implementing a form of Bayesian reasoning, i.e., Bayesian filtering, for a certain goal.

## 2.2 Hybrid augmentation

Hybrid augmentation includes research on autonomous systems whose perceptual, cognitive or motor capabilities are extended via the integration with non-autonomous parts, e.g., neurons connected to robots can gain mobility, humans connected to virtual reality setups can explore new worlds and access perceptual abilities otherwise impossible.

A major source of inspiration for this line of research can be found in classical examples of sensory substitution in humans. Bach-y-Rita et al. (1972) for instance introduced a system converting visual inputs captured by a camera into tactile signals on the back of different subjects, showing that blind subjects learnt how to "visualise" this type of haptic information. More recent devices use for example sound (Striem-Amit et al., 2012; Pasqualotto & Esenkaya 2016) and electrical stimuli on the tongue to replace visual sensations (Danilov & Tyler 2005; Friberg et al., 2011).

A minimalist approach for sensory substitution, showing the flexibility and potential of this paradigm, has been proposed by Froese et al. (2011), with the use of the "enactive torch", a simple device whose vibration intensity is correlated with the distance between a user and objects in the environment measured with an infrared sensor. While the sensory information provided by this device is significantly limited compared to work such Bach-y-Rita et al. (1972), people controlling an enactive torch appear to learn to identify the shape of different objects over time. In the context of our overview of hybrid augmentation, sensory substitution devices shed light on the role of neuroplasticity as a mechanism enabling not only the replacement of a lost sensory modality, but also the acquisition of new ones via "sensory augmentation" (Froese et al., 2011).

---

[1] The Shannon mutual information between past and future of random variables.

More recent examples of sensory augmentation have emerged following developments of deep learning methods, with studies of perception relying on artificial perceptual systems built using artificial neural networks. *Artificial Perception* (Sinapayen et al., 2021)[2] in particular showcases work on artificial illusions falling squarely in our definition of hybrid augmentation. Illusions are percepts creating a contradiction between different perceptions of a system in an observer, or between the perception of that system and what the observer knows about it: for example, we know that the "Rotating Snakes Illusion" (Kitaoka, 2003) is a static image, but we somehow perceive it as moving. To investigate the principles behind visual motion illusions, Sinapayen & Watanabe (2021) used an artificial neural network to generate novel visual illusions, creating perceptions of motion that need not be available in naturalistic settings. This approach can be used in psychophysical experiments to develop and study potentially unknown aspects of perceptual experience, extending work such as Kobayashi et al. (2022) where other artificially created illusions were used.

Suzuki et al. (2017) similarly investigate phenomenological characteristics of visual hallucinations, rather than illusions, in humans. Visual hallucinations are referred to as false perception in the absence of the physical counterparts during an awakening state, thus being distinguished from both visual illusions and dreams. Using visualisation techniques consistent with computational processing attributed to the human brain under visual hallucinations, Suzuki et al. (2017)'s "Hallucination Machine", an artificial neural network trained with Google's Deep Dream algorithm, successfully simulated marked phenomenology often reported in visual hallucinations induced by psychedelics. With the support of virtual reality technology, these artificial hallucinations were then displayed in real time to subjects whose reports indicated strong changes in their ability to perceive the world, as in the case of the "dog world" in Fig. 1.

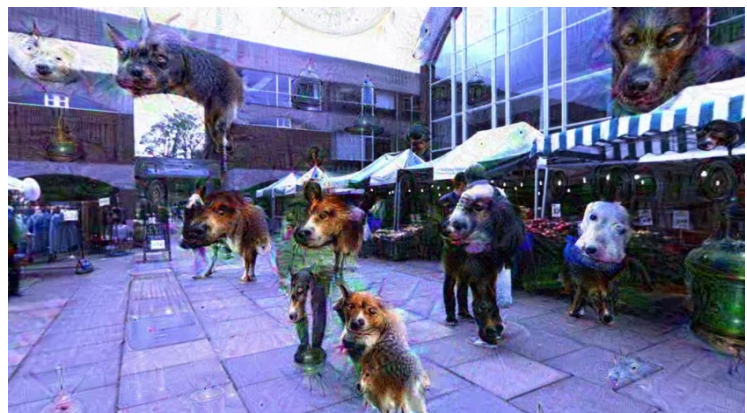

**Figure 1. The Hallucination Machine.** The world as seen with a head-mounted virtual reality display replicating visual hallucinations in a "dog-world": a world where dog features, such as faces, eyes, etc. are extrapolated in all possible existing patterns appearing in an immersive virtual reality environment (figure from Suzuki et al. (2017), CC BY 4.0).

---

[2] Artificial Perception (Sinapayen et al., 2021) appeared as a separate special session at the Artificial Life conference in 2021-2022, led by one of the co-authors.

When it comes to motor augmentation on the other hand, a prominent area of research is centred around studies of *cyborgs*, i.e., cybernetic organisms, part biological and part machine, with the main organisms ranging from insects and rodents to plants and microorganisms, or even humans. A cyborg is usually distinguished from other "mixed" systems because its artificial components can not only replace missing parts of the organism (e.g., like prosthetic limbs in humans), but also extend its original capabilities. Ando & Kanzaki (2020) classified cyborgs into two categories: 1) those that implement machines on biological organisms and 2) those that use biological parts to augment robots. In the first group, we find for example attempts to externally control insects by mounting electronic components on them, such as in the case of the cyborg cockroach (Kakei et al., 2022). The second category includes works using for instance insect antennae as pheromone sensors on mobile robots (Kuwana et al., 1995; Kuwana et al., 1999).

Other than in insects, the integration between artificial and biological systems has been an active research area, interestingly, in the case of plants. Plants are often treated as closer to inanimate objects than to living beings, even when kept as houseplants, partly because of the slow timescale of their reactions compared to humans and their pets. Several augmentation projects improve this perception by endowing plants with robotic bodies that accelerate their reaction time or increase their agency on the world. In Demaray et al. (2015)'s "Floraborgs" and Sareen & Tiao (2018)' "Cyborg Botany" for instance, potted plants are augmented with robotic wheels to move towards sunny spots and increase their growth rate, see Fig. 2, while Umoz, a hexapod robot that acts as a host for moss, adapts its behaviour depending on the needs of the moss species, e.g., sunlight, water (MTRL, 2021).

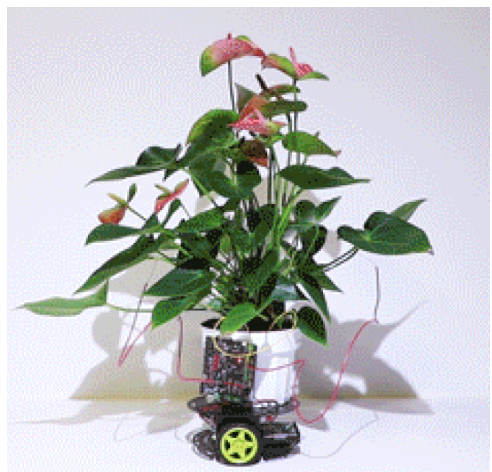

**Figure 2. Elowan/Plant-Robot Hybrid.** An example of a plant endowed with a wheeled robotic body that can be used to follow a light (credit: Harpreet Sareen, CC BY 4.0, figure adapted from Sareen & Tiao (2018)).

Mirroring the common perception of plants as autonomous systems with limited cognitive presence, in-vitro neurons can be seen as a common basis for complex cognition without much autonomy (at least without a body). Masumori et al. (2018) augment in-vitro rat neurons, embodying them into a wheeled robot with sensors and actuators (see also Shahaf & Christakis, 2001; Tessadori et al., 2012; Aaser et al., 2017 for earlier related work, or Kagan et al., 2022 for a recent partial replication). These neurons are then able to learn and perform a navigation task based on the sensorimotor mapping between the robot and the cultured neurons.

### 2.3 Hybrid interaction

In the previous section we looked at the integration of an autonomous system with non-autonomous parts, reviewing areas of the literature that show how hybrid augmentation can lead to systems with extended or even new perceptual and motor capabilities. With a stronger commitment to autonomy, here defined following Di Paolo & Iizuka (2008) as the self-sustaining process that leads to the emergence of an individual and its *identity* under precarious conditions, we can similarly observe that different autonomous systems interacting in non-trivial ways can promote to the emergence of new dynamics and behaviours.

Our hybrid approach stresses the importance of active interactions over passive ones. Rather than observing a murmuration of starlings to understand its behaviour, we believe that building a robot starling that can directly interact with the murmuration leads to a better understanding of collective interactions. A similar distinction appears under the name of "replay" or "playback experiments" in different areas of biology and neuroscience, showing the importance of active interactions to better explain the behaviour of different living organisms (see, e.g., Buckley & Toyoizumi, 2018 and references therein). These experiments involve pre-recording the behaviour of an organism and testing how a different target biological organism behaves when it tries to interact with an "artificial" system, i.e., the recorded behaviour. In other cases, artificial systems appear as synthetic recordings generated from a mathematical model built using the organism's behaviour, thus having a target organism interacting with an entirely synthetic, or simulated system. Matsunaga & Watanabe (2012) for example used "virtual plankton" to study what kind of moving behaviour in plankton can more often elicit predatory behaviour in medaka, showing that movements characterised as pink, or 1/f, noise appear to induce the most responses.

This type of approach is also used in studies of "mixed" or "hybrid societies" (Halloy et al., 2007; Schmickl et al., 2013; Hamann et al., 2016; Bonnet et al., 2019; Landgraf et al., 2021). These systems include different classes of individuals interacting in nontrivial ways, giving rise to complex behaviours comparable to those of a "society". Halloy et al. (2007) in particular introduced a chemically dressed robot (i.e., a robot bearing the correct chemical signal) into a cockroach population and showed that decision making in a mixed cockroach population can be influenced by the robot. Inspired by both Halloy's work and by a large number of other experiments studying hybrid interactions in fish (see Fig. 3 for a standard setup and Cazenille et al., 2018 for a review), Cazenille et al. (2018) instead placed a robot fish within a group of zebrafish to form a hybrid school. In their experiments, they showed that groups containing the robot fish could generate collective behaviours comparable to that of groups composed solely of living fish, measuring the impact of the morphology and movements of the robot fish on the dynamics of the hybrid school. Iizuka et al. (2019) used a similar setup with medaka (Oryzias Latipes), replacing the robotic fish with a computer-generated one that was displayed on a monitor attached to the side of the fish tank. The movements of the biological fish were captured in real time, while the artificial agent's movements were mediated by the movements of the real fish: following a standard Boids rule computed from the locations of both the biological and the artificial fish. Their work showed results consistent with the literature, emphasising the richer behaviour of biological systems that actively interact with artificial ones, even for fish (medaka) simply looking at a screen. Shirado and Christakis (2017) on the other hand included bots (artificial programs pretending to be people) as players in a group of people participating in a coordination game. In this hybrid setup, the bots influenced the behaviour of human players, resolving conflicts that would normally be difficult to tackle in a society consisting only of human players.

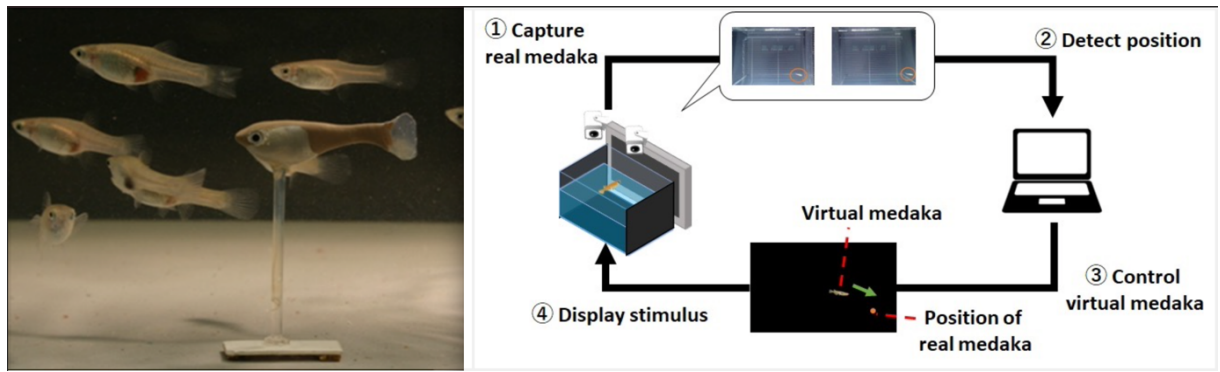

**Figure 3. Robot-biological fish hybrid interaction.** In the left panel, a typical example of robot-fish interaction (figure adapted from Bierbach et al., 2018, CC BY 4.0), while on the right one we have a schematic of a setup for closed-loop hybrid interaction, here using a simulated fish (figure from Iizuka et al., 2019, CC BY 4.0). A standard hybrid interaction includes a biological organism (phase 1, e.g., a fish), whose features are detected (phase 2, e.g., movement are recorded) to generate control signals for an artificial system (phase 3, e.g., a computer-generated or robotic fish) that can be seen by, and thus interacts with the living organism (phase 4).

Another interesting example appears in Hamann et al. (2015), where robots and plants, portrayed in a symbiotic society named "flora robotica", together produce new kinds of architectures. Importantly, in this setup robots and plants are proposed to be treated on equal footing: as (different) classes of autonomous individuals that can exchange different kinds of information and energy: chemical, mechanical, etc. The crucial aspect of the self-organising behaviour of this hybrid society lies in the fact that robots and plants are both allowed to grow: plants grow while supported by mechanical structures constituting the robots' bodies, with the robots also controlling light, water, temperature, and other variables for the plants; robots "grow" either by prompting experimenters, asking for their active interventions, i.e., the addition of new components (Hamann et al., 2015), or by self-assembling new parts, e.g., using on-site scaffolding structures that can be updated over time (Hamann et al., 2017).

Another example of the hybrid interaction approach appears in a behavioural version of the Turing test. The Turing test, proposed by Turing as an "imitation game", is commonly understood as measuring, or providing a proxy for artificial intelligence. While the original idea involves interactions through chat conversations, a behavioural version of the test can be implemented using variations of the original setup such as the handshake Turing test (Avraham et al., 2012) or the perceptual crossing experiments for agency detection (Auvray et al, 2009; Iizuka et al., 2012; Froese et al, 2014). In the handshake test, human participants interact with a teleoperated robotic interface producing handshake-like behaviour and are later asked to report on what kind of movements, produced while interacting with the interface, felt more natural. The perceptual crossing task involves an even simpler setup, based on an experiment where participants need to distinguish a human interacting partner from non-human moving objects. To do so, participants can use only a cursor that they can slide along a line to perceive a tactile stimulus when encountering something (human or not). In this paradigm, players have been shown to find human partners by generating mutually, temporally sustained crossing motions, which can be used to distinguish humans from objects that are represented by either static or "shadow images" of the other player (see Auvray et al, 2009; Froese et al, 2014), or by non-interactive replays from previous trials (Iizuka et al., 2012).

## 3. SPECIAL SESSIONS 2018-2022 - A REPORT

In the second part of this overview, we turn our attention to a series of outputs of Hybrid Life sessions at the International Conference on Artificial Life, where Hybrid Life has featured as a special session between 2018 and 2022. Over these five years, Hybrid Life sessions have contributed to the publication of over 30 conference proceedings[3], while also featuring 2 presentations by internationally renowned invited speakers and the introduction of 1 competition open to participants not only of our sessions, but from around the world. This output will be organised following the same three themes described in Section 2, see Table 1. for a thematic overview.

| Theme | Focus | Works |
|---|---|---|
| Theories of systems | General principles | Noguchi et al. (2018), Aguilera et al. (2019), Berry & Valero-Cuevas (2020), Man & Damasio (2020), Damiano & Stano (2020), Sinapayen & Witkowski (2020)*, Mellmann et al. (2020), Ovalle (2021), Segerståhl (2022) |
| | FEP - Predictive processing | Baltieri & Buckley (2018; 2019), Cowley et al. (2018), Baltieri et al. (2020) |
| | IIT | Aguilera & Di Paolo (2018), Rodriguez (2022) |
| Hybrid augmentation | Augmented machines/robots | Masumori et al. (2018), Knight et al. (2019); Ofner & Stober (2019), Sandamirskaya et al. (2022)** |
| | Augmented organisms | Froese et al. (2018), Suzuki et al. (2018), Suzuki et al. (2019b), Huang et al. (2021)***, Eden et al. (2022)*** |
| Hybrid interaction | Human - robot/avatar/machine interactions | Dotov & Froese (2020), Matthews & Bongard (2020), Maruyama et al. (2021), Tsuruta et al. (2021), Suda & Oka (2021), Rockback et al. (2022) |
| | Non-human organisms - robot/machine interactions | Mariano et al. (2018), Rampioni et al. (2018), Kadish et al. (2019), Chen et al. (2019), Dorin et al. (2018), Iizuka et al. (2018; 2019), Luthra & Todd (2021) |

**Table 1. Outputs of Hybrid Life special sessions.** Conference proceedings, competitions and invited talks appearing at Hybrid Life sessions hosted by the International Conference on Artificial Life, 2018-2022.
Notes: * A competition presented by some of the co-authors at Hybrid Life 2020. ** Work presented by an invited keynote speaker at the Hybrid Life session in 2021, not submitted to the session. *** Work presented by an invited keynote speaker at the Hybrid Life session in 2022, not submitted to the session.

### 3.1 Special sessions - Theories of systems and agents

While originally introduced as a theory of consciousness, Integrated Information theory (IIT) has more recently been proposed to provide a measure of integration of systems and agents (see Section 2.1). Aguilera & Di Paolo (2018) extends some of this work by using an approximation of the IIT measure, *phi* (originally proposed to measure consciousness) to study autonomy in possibly infinitely large Ising models. In the context of studies of autonomy, Aguilera & Di Paolo (2018) propose that their approximated phi measure could be used to distinguish cases when a system can be partitioned into an interacting agent-environment pair, or when it should instead be treated as a whole. Similarly, work by Rodriguez (2022), provides a complementary view that attempts to further connect work in IIT to the notion of *enactive* autonomy (Varela et al., 1991), using gliders and their cognitive domain (Beer, 2014) as a toy model. Another perspective on autonomy appears in Ovalle (2021), where it is linked to an agent's intentionality and normative behaviours[4], and contrasted to approaches such as reinforcement learning, more often relying on externally assigned cost functions.

---
[3] Proceedings include both full manuscripts based on unpublished work, and extended abstracts containing preliminary results or reports summarising relevant manuscripts published elsewhere.

Different aspects of normativity and autonomy are also generally explored within active inference, a process theory inspired by the free energy principle (FEP, see section 2.1), including the relation between homeostatic stability and surprisal minimisation, and how normative constraints can be acquired or learnt over time. Cowley et al. (2018) focus on mechanisms of attention that are defined in different models of predictive coding to describe how systems can attribute value to different inputs (Keller & Mrsic-Flogel, 2018), both inspired by and independent of the FEP. Cowley et al. (2018) in particular use a model proposed as a biological plausible implementation of activity in the neocortex, Hierarchical Temporal Memory (Hawkins & Blakeslee, 2004). Cowley et al. (2018) implement attention via the modulation of units whose activation reflects correct predictions from a previous time step. This model paves the way for extensions of the original Hierarchical Temporal Memory architecture to temporal prediction models in different biological and artificial systems, and is an important step in comparing different predictive coding implementations.

Using active inference more specifically, Baltieri & Buckley (2018) discuss different perspectives on cognitive architectures based on the presence, or lack, of functional modularity. This work argues that functional modularity, understood as the sequential, functional partition and information encapsulation of perception, cognition and action (see also "sense-plan-act" models, Brooks, 1991) originates from the "separation principle" in control theory. Under this principle, estimation of the latent states of the world and control of the system based on our best estimate of its hidden states can be tackled separately, albeit under some strong assumptions. In cognitive science, this is often used to justify how perception, in the form of state estimation/prediction (Keller and Mrsic-FLogel, 2018) and motor control (Kawato, 1999) can be seen as separate "modules" in a cognitive architecture, see also Mellmann et al. (2020) and references therein for applications in robotics. Baltieri & Buckley (2018) contrast models requiring an "efference copy" mechanism (Kawato, 1999), i.e., a copy of motor signals to be known by a system, from frameworks not requiring explicit knowledge of motor commands, such as active inference (Friston, 2010). Active inference replaces the more widely adopted efference copy with inferences about motor commands themselves, subjected to the presence of proprioceptive information (Friston, 2010). Proprioception is the sense of self-movement, body force and position, explaining how we normally know where different parts of our body are, and in what position they are, at all times. Berry & Valero-Cuevas (2020) use proprioception to argue that physiological proprioceptive afferents contribute to the formation of body representations, to neuromuscular control, and potentially even to the emergence of the sense of agency and self. Instead of seeing proprioceptive and motor signals as separate, the authors present an extended concept of "active sensing" seeing them as dynamically blending parts of the same motor-perceptual continuum. They call it "sensory-motor gestalt", extending classical accounts of gestalt found in visual perception to the motor domain using various perspectives from biomechanics, differential geometry, and physiology to understand the emergence of body representations and the self in the context of proprioception and motor actions in the physical world.

While considering different kinds of representations, Noguchi et al. (2018) instead show mechanisms that can contribute to the formation of stable patterns of neural activity by demonstrating how the complexity of an agent's movement patterns, or behaviours, can generate place cell-like structures within a recurrent neural network. On the other hand, Man & Damasio (2020) emphasises how not only external information can affect the internal state of a system, but how the reverse ought to be also true to establish a counterfactual concept of "meaning" through a causal relationship, where the performance on an external task is considered to be desirable for a system if evaluations of the effects of a proposed solution lead the system to maintain its internal variables within bounds, i.e., homeostasis.

---

[4] Roughly, normativity implies that behaviours are "good" for an agent from its own perspective, for example goal-directed behaviours conducive to its own survival, with goals intrinsically assigned by the agent. Goals and norms are in some special cases used interchangeably (Barandiaran et al., 2009) but should not be generally confused, hence our use of "normative".

A regular criticism of predictive coding and the FEP hinges on the possible lack of "meaning" and normativity: what would prevent a system simply minimising prediction error from finding a dark room where observations are more easily predictable (i.e., all the same, black)? Baltieri & Buckley (2019) clarifies that this question only makes sense when a system's goal, e.g., surviving as long as possible, is confused with how such a goal can be achieved, e.g., minimising prediction error, (see Sun et al., 2020 for an example of this confusion). Describing systems as inference machines can be misleading if this algorithmic view is removed from studies of normativity in embodied agents, living systems and robots that can be used to constrain problems of this kind and remove this (only apparent) paradox. Other worries however still exist for the FEP and active inference, especially about their precise claims. While initially proposed as a theory of brains (Friston, 2010), it has since then been used to describe the origins of all known life (Friston, 2013, but see Biehl et al., 2021), and more recently, it has further been introduced as a theory of systems (Friston, 2019, but see Bruineberg et al., 2021). Baltieri et al. (2020) unpacks some of these ideas, showing how the process theory, active inference, can be used to model a number of different systems (cognitive or not, living or not), without necessarily having to invoke the more general principle, the FEP. In this paper, the authors derive a set of equations of an action-perception loop by treating a steam engine-Watt governor coupled system as an agent-environment model, describing beliefs, goals and predictions about the behaviour of the system. In this somewhat paradoxical set up, the authors discuss the merits (or lack of) of an overarching theory that can encompass different classes of systems, and the possible outcomes of a "pan-inferential" interpretation of behaviour as the starting point for a theory of agency and systems.

Aguilera et al. (2019) propose a quantitative account of affordances, defined as "directly perceived environmental possibilities for action" in agents, a central construct of ecological psychology and a defining feature of agency within this field. Aguilera et al. (2019) provide an analysis of the common background between the definition of affordances, using examples and experimental results in ecological psychology, and *empowerment*[5] (Klyubin et al., 2005), whose maximisation roughly corresponds to the idea of an agent seeking to maximise its options to act in the future. This work constitutes a step in the formalisation of affordances using tools from information theory with an application to theories of agency and intrinsic motivation using empowerment, while also attempting to connect empowerment, a quantity notoriously difficult to calculate in practice, to a vast literature of quantitative results in ecological psychology, with the goal of providing a proxy for its experimental validation.

Inspired by more fundamental attempts to define life and biological principles, work by Damiano & Stano, (2020) propose a new route for the study of AI systems using models from synthetic biology to complement existing robotic and digital implementations. On the other hand, Segerståhl (2022) proposes a unifying theory of systems inspired by the biological cell. In particular, Segerståhl (2022) introduces a model that can provide part of the necessary components to bridge the gap between autopoietic accounts of the living cell (Maturana and Varela, 1980) and the (M,R)-systems (Rosen, 1991), using notions of information processing and a diagrammatic language formalising some core ideas from biology. As a way to challenge these and the ever-so increasing number of theoretical proposals for theories of systems, and life more specifically, Sinapayen & Witkowski (2020) introduced a competition with the goal of identifying living systems from non-living ones. The main goal was to sort unlabelled 2D trajectories of living (birds, spiders, etc) and artificial systems (robots, artificial chemistries, etc) into "living" and "non-living" categories using theory-based algorithms (AI/black-box learning algorithms were forbidden). The winning algorithm turned out to be based on the idea that living systems appear to behave in a scale-free manner, i.e., their behaviour can be described by power laws that characterise fractal structure, i.e., (self-)similar at different spatiotemporal scales.

---

[5] Empowerment is defined as the channel capacity between an agent's actuators at time t and its sensors at time t+1.

### 3.2 Special sessions - Hybrid Augmentation

Ideas inspired by the theoretical frameworks described above (FEP, IIT, etc.) and, in general, theories of systems have percolated through a series of applications in systems augmentation. For instance, Ofner & Stober (2019) examine the ability of machine learning models to integrate models from neuroscience, in particular predictive coding and active inference (Friston, 2010). Their model mimics multi-modal exteroception and interoception in the human brain by learning how to make predictions about the future using visual stimuli in the form of videos and electroencephalography, electrooculography and electromyography data recorded in human subjects watching the same videos.

On the other hand, Sandamirskaya et al. (2022)[6], show applications of systems augmentation in robotics and autonomous systems where neuromorphic hardware is used to run bio-inspired artificial neural networks. Neuromorphic devices are an energy efficient, low-latency alternative to more standard hardware, displaying capabilities similar to biological systems on a number of different tasks, including cases requiring some form of cognitive abilities such as learning, motor control and perception. Using in particular spiking neural networks, a biologically plausible and particularly efficient implementation of artificial neural networks on neuromorphic hardware designed for asynchronous, event-based (spike) computation, Sandamirskaya et al. (2022) show that applications of these networks to navigation (path integration, SLAM), learning (supervised and unsupervised) and cognitive architectures can lead to a new path for studies of intelligence bridging the gap between biological and artificial intelligence paradigms. Knight et al. (2019) implement a similar setup, relying however on more standard rate-based artificial neural networks, trained on visual data from specified routes, with the goal of building a "familiarity-based" navigation algorithm integrated on robots. The network can be trained to output a "familiarity" score for any visual input, which can be used to control a robot whose task is to return to its initial position by choosing routes that look the most familiar. Masumori et al. (2018) use a similar setup, reporting on experimental and simulation results of, in this case, biological neural networks controlling simple robots, see Fig. 4, showing that the networks tend to cut off neurons that relay "uncontrollable" input from the environment. By comparing the spiking dynamics in networks facing a solvable task and networks facing a difficult or unsolvable task, the authors show that the networks either manage to learn the task (therefore decreasing the amount of incoming input), or after failing to learn the task, cut off input neurons, therefore decreasing the amount of incoming input by redefining the boundary between self (connected neurons) and non-self (environment and disconnected neurons).

---

[6] Sandamirskaya was invited as a keynote speaker for the Hybrid Life special session in 2021.

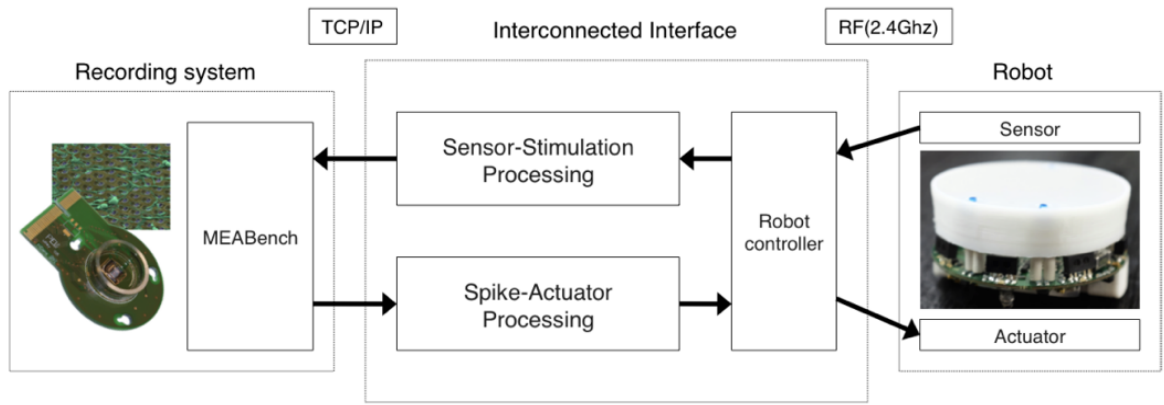

**Figure 4. Cultured neurons controlling a robot.** Rat neurons are used in this experiment to relay neuronal activity to a wheeled robot trying to avoid hitting walls and to process sensory values from the environment captured via the robot's sensors in a closed loop (figure from Masumori et al., 2015, CC BY 4.0).

While these works show the potential of integrating biologically inspired methods or even parts of living systems into robots, in recent years new technologies have also driven the development of augmentation techniques for biological systems using artificial components, in some cases leading to new forms of cognitive processing, as discussed in Section 2.2. Froese et al. (2018) for instance introduced a synthetic model of the origins of the genetic code based on an iterated learning approach originally developed as a model of language evolution. The authors trained a multi-layer perceptron network as a coding system from codons to the 11-dimensional amino acid feature space. While iterating the learning algorithm, 'donor' protocells transferred subparts of their codon assignment to 'receiver' protocells, which in turn adjusted their coding system to more closely align with the 'donor' protocells from which they received codon assignments. They found the emerging artificial genetic code significantly overlaps with the one found in the literature even without modelling detailed chemical pathways.

Huang et al. (2021) and Eden et al. (2022)[7]'s works provide outstanding examples of biological system's augmentation for motor control, focusing on both existing movement abilities and the development of new degrees of freedom. The former includes work on power augmentation, such as exoskeletons, workspace augmentation, such as teleoperating robotic limbs, and command augmentation, where devices are used to address possible limitations of the motor system, e.g., active noise cancellation for tremors. The latter, on the other hand, involves work on augmentation by extension, such as a third robotic hand or a simulated cursor on a screen controlled by a brain-machine interface (either invasive, e.g., with ECoG, or not, e.g., EMG), by transfer, such as with feet controlling robotic arms while sitting (Huang et al., 2021), and others, see Eden et al. (2022) for an extensive review.

Similarly to Suzuki et al., 2018's "Hallucination Machine", in which people can experience synthesised visual hallucinations while wearing a head mounted display (see Section 2.2), Suzuki et al. (2019b, 2019a) use immersive virtual reality to investigate how sensorimotor contingencies influence visual awareness in humans. Suzuki et al. (2019a, b) in particular developed a virtual reality setup combined with motion tracking, allowing participants to interact with novel-shaped, virtual 3D objects. This work found evidence that sensorimotor contingencies can influence the process of visual perception for unfamiliar 3D virtual objects, or in other words that the ability to manipulate objects affects our visual experience of them, showing how the use of virtual reality technology can shed light on experimental psychology and studies of perception, and in the future even extend our very own perceptual experience.

---

[7] Huang and Eden were invited as keynote speakers for the Hybrid Life special session in 2022.

### 3.3 Special sessions - Hybrid Interactions

In our definition of hybrid interactions, we required either 1) a high level of autonomy in all interacting parts of a whole (e.g., humans using exoskeleton are not to be included here), 2) distinguishable classes of individuals for heterogeneous interaction (e.g., a colony of ants is generally homogenous), or ideally both. This normally would exclude a large number of studies in swarm or collective intelligence, however an example of how swarms can be seen in light of our hybrid interactions proposal can be found in Rockbach et al. (2022). Rockbach et al. (2022) focus on simulations of the interactions between a human and a robot swarm treated as a macro-scale individual, with applications in the context of search-and-rescue activities where unmanned vehicles can be used cooperatively with humans for on-field operations. A similar line of research, steered towards the study of autonomy and the interaction of multiple goals on interactive systems emerges also in work by Dotov & Froese (2020), where the authors propose the concept of "dynamic interactive AI" as a hybrid interaction of human and artificial systems. Dynamic interactive AIs are proposed as systems self-organising through dynamic interactions with humans. Using the authors' own research on human-machine interaction as an example, they contrast the differences between this approach and non-interactive AI, highlighting the advantages of a shift towards an interaction-based learning paradigm using chaotic dynamical systems as an example. Interestingly, this approach has led to applications in gait cadence regulation for patients with motor disorders, i.e., the regulation of walking patterns and their related speed in patients affected by Parkinson's disease. Dotov et al. (2019) showed how the active interaction of humans and artificial systems displaying some form of autonomy (with the presence of internal dynamics not directly manipulated by external users) produces better entrainment, i.e., synchronisation of auditory cues and gait, helping patients with motor disorders to control their gait cadence.

Working in a different domain, studying the foundations of human-robot interactions, Maruyama et al. (2021) reach similar conclusions about the synchrony between humans and artificial agents while exploring the mimetic behaviour of a humanoid robot called Alter3. Alter3 can mimic the motion of humans appearing in its visual field, while also storing different motion sequences that can be used to generate intrinsically driven movements based on an internal simulator. Altogether, this repertoire of possible behaviours gives rise to interactions used to investigate the social components of information exchange in human-based and potentially hybrid societies. Using measures of directed information flow, the study observes that when Alter3 fails to imitate human motion, humans tend to mimic Alter3 instead.

Matthews & Bongard (2020) also look at human-robot interactions, focusing however on simulated robots and interaction via natural language, rather than visual input and bodily movements, in a human-machine collaborative design paradigm. In this work the authors propose a mechanism of "crowd grounding", extending ideas from action grounding (how words could be translated into actions) and reward grounding (how novel rewards could be generated given past policy/reward pairs) into an approach where active interactions with humans are mediated through an online messaging platform. Tsuruta et al. (2021) study a somewhat complementary case, asking whether virtual agents can be designed to facilitate the communication between human agents interacting on an online platform. Using an agent-based model, they show evidence that artificial systems can in principle be used to improve communication in a hybrid interacting system, by broadcasting "attractive" information to invite other individuals to join a cluster, by transferring any kind of information across existing clusters or with a combination of these two strategies. Suda & Oka (2021) also study interactions in humans and virtual agents, here however focusing on pairwise interactions and with virtual agents appearing as avatars representing humans. This work sets off to study differences that could appear in human-avatar interactions as opposed to an avatar-avatar scenario common in different modern online platforms. They also ask whether interaction, with avatars mimicking a subject's facial expressions, can influence human decision making, showing evidence that mimicry appears to improve cooperative behaviour.

Looking at hybrid interactions involving also biological systems other than humans, Kadish et al. (2019) propose to investigate niche differentiation, i.e., how competing species use the environment in different ways to be able to coexist, in hybrid societies using soundscape ecology, the study of acoustic relationships between organisms. Luthra & Todd (2021) focus instead on social search, i.e., the use of social information to locate resources, paving the way for a formal study of some of the possible mechanisms leading to information sharing in, among others, hybrid interacting systems (e.g., odours in Halloy et al., 2007). Chen et al. (2019) on the other hand describe a robotic interface that can facilitate interactions between humans and a microbial ecosystem. Using a wide array of sensors, their robot monitors the state of a traditional Japanese pickle fermentation bed (called "nukadoko" in Japanese) containing colonies of lactic acid bacteria, yeasts, and gram-negative bacteria. The interface, acting as an artificial personification of the colonies, alerts the human if the nukadoko needs stirring, and provides information about one of the three fermentation stages and the predicted taste of the pickles.

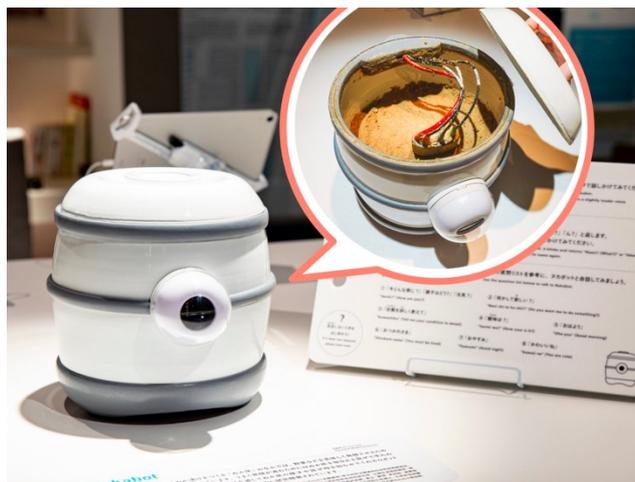

**Figure 5: Nukabot.** A robotic interface for a microbial ecosystem, a traditional Japanese pickle fermentation bed. (Photo Credit: The National Museum of Emerging Science and Innovation, used with permission).

Mariano et al. (2018) on the other hand deployed robots to develop forms of communication with honeybees. In nature, bees are known to use vibration patterns to send signals to other bees, asking them for example to freeze or slow down. In this paper, using robotic agents, different vibration patterns were evolved using genetic algorithms to function as a language between artificial systems and bees, in this case persuading them to stop moving in the vicinity of the buzzing robots. Similarly, Rampioni et al. (2018)'s proposal also aims to build a new language, in this case a chemical one, between synthetic and biological cells (a pathogenic bacterium, Pseudomonas aeruginosa) taking advantage of the way systems can learn to interact over time. Using once again bees, Dorin et al. (2018) instead created a system generating planting arrangements for greenhouses that can maximise pollination in what they promisingly dubbed a *techno-ecological system* following the idea of having humans actively cooperating with artificial agents to design and better control complex systems.

Under the same hybrid interactions paradigm, Iizuka et al. (2019) performed experiments with a biological organism, a fish, and an artificial agent, as already discussed in Section 2.3. Interestingly, this paper also describes the potential to test different theories of interaction by performing other experiments with biological organisms and artificial agents. Previous results from Iizuka et al. (2018) for instance describe how a neural network based controller, trained to learn the sensorimotor mapping of a fish from biological data, can outperform the Boids rule used in Iizuka et al. (2019). Iizuka et al. (2018) however tested their network-based controllers on individuals part of purely synthetic, i.e., simulated, swarms. Further investigations will thus be necessary to validate this result for the kind of hybrid interactions with living systems discussed in Iizuka et al. (2019).

## 4 CONCLUSIONS

Artificial life (ALife) is a research field that has traditionally included work investigating "life as it could be" (Langton, 1989). In this overview, we introduced an active area of research that has recently emerged at the intersection of ALife and other fields including psychology, human- and brain-machine interfaces, mathematics, engineering, virtual and augmented reality technologies. This area, named by the authors Hybrid Life, focuses on three complementary themes: 1) theories of systems, studying formal definitions of systems and agents, 2) hybrid augmentation, asking how a system's perceptual, cognitive and motor capabilities can be extended using components of different kinds, and 3) hybrid interaction, exploring the cooperation of systems with high degrees of autonomy with a focus on interaction between individuals of different classes (e.g., biological and artificial). In the first part of this overview, we focused on a number of background studies that we believe constitute either foundational background to trace the origins of three themes, or prominent attempts to fully implement core ideas of Hybrid Life. In the second part we discussed a series of examples that appeared in a series of Hybrid Life special sessions organised by the authors as part of the annual International Conference on Artificial Life over a period of 5 years (2018-2022). While covering all the contributed material to the special sessions, several works in different relevant areas couldn't be explored in detail in the first part. Some of these works include discussions about ethics, and in our specific case the ethics of ALife systems. Most of the technologies and frameworks described in this manuscript, including the papers published in Hybrid Life sessions, have potentially limitless applications, however their ethical implications and risks deserve more careful consideration. Insects whose movements are manipulated by miniaturised radio-controlled components could be used not only for agricultural purposes by transporting, e.g., pollen, for rescue and exploration, but also as weapons of war (Siljak et al., 2022). Virtual reality technologies can create entire new worlds to explore, but might pose a risk to, e.g., child development (Kaimara et al., 2022). Following these and several other overt examples, a major focus of Hybrid Life research in the future should thus include an exploration of the ethical commitments and moral assessments and obligations towards the development of theories of systems, their augmentations and interactions (Witkowski & Schwitzgebel, 2022).